\pdfoutput=1
\documentclass[11pt,a4paper]{article}
\usepackage[hyperref]{acl2019}
\usepackage{times}
 \usepackage{todonotes}
\usepackage{latexsym}
\usepackage{amsmath}
\usepackage{amssymb}
\usepackage{url}
\usepackage[noend]{algorithm2e}
\usepackage{graphicx}
\usepackage{tikz-qtree}
\usepackage{color}
\usepackage{xcolor,colortbl}
\usepackage{setspace}
\usepackage{pgfplots}
\usepackage{float}
\usepackage{amsthm}

\definecolor{Gray}{gray}{0.85}
\definecolor{lGray}{gray}{0.95}
\newcolumntype{a}{>{\columncolor{lGray}}r}
\newcolumntype{b}{>{\columncolor{lGray}}r}
\newcommand{\mina}{\texttt{MINA}\xspace}

\aclfinalcopy 

\newtheorem{example}{Example}

\mathchardef\mhyphen="2D
\newcommand{\GG}[1]{}

\newcommand{\squishlist}{
 \begin{list}{$\bullet$}
  { \setlength{\itemsep}{0pt}
     \setlength{\parsep}{3pt}
     \setlength{\topsep}{3pt}
     \setlength{\partopsep}{0pt}
     \setlength{\leftmargin}{1.5em}
     \setlength{\labelwidth}{1em}
     \setlength{\labelsep}{0.5em} } }

\newcommand{\squishend}{
  \end{list}  }

{\begin{myalgo}[#1]
\centering
\begin{minipage}{#2}
\begin{algorithm}[H]}%
{\end{algorithm}
\end{minipage}
\end{myalgo}}

\title{Using Automatically Extracted Minimum Spans to Disentangle Coreference Evaluation from Boundary Detection}

\author{Nafise Sadat Moosavi$^{1}$ \and Leo Born$^2$ \and Massimo Poesio$^3$ \and Michael Strube$^{4}$
       \\
       $^1$Ubiquitous Knowledge Processing (UKP) Lab, Technische Universit\"at Darmstadt\\
       $^2$Institute for Computational Linguistics, Heidelberg University\\
	   $^3$School of Electronic Engineering and Computer Science, Queen Mary University of London\\
       $^4$Heidelberg Institute for Theoretical Studies gGmbH\\}

\date{}

\begin{document}
\maketitle
\begin{abstract}
The common practice in coreference resolution is to identify and evaluate the maximum span of mentions.
The use of maximum spans tangles
coreference evaluation with the challenges of mention boundary detection like prepositional phrase attachment.
To address this problem, minimum spans are manually annotated in smaller corpora.
However, this additional annotation is costly and therefore, this solution does not scale to large corpora.
In this paper, we propose the \mina algorithm for automatically extracting minimum spans to benefit from minimum span evaluation in all corpora.
We show that the extracted minimum spans by \mina are consistent with those that are manually annotated by experts.
Our experiments show that using minimum spans is in particular important in cross-dataset coreference evaluation, in which detected mention boundaries are noisier due to domain shift. 
We will integrate \mina into \url{https://github.com/ns-moosavi/coval} for reporting standard coreference scores based on both maximum and automatically detected minimum spans.
\end{abstract}

\section{Introduction}
\label{sect:intro}
Coreference resolution is the task of finding different expressions that refer to the same real-world entity. Each referring expression is called a mention.
The common approach to annotate coreferring mentions is to specify the largest span of each mention.
The problem with using maximum spans in coreference evaluation is that a single mention may have different maximum boundaries based on gold vs. automatically detected syntactic structures.
For instance, variations in prepositional phrase attachment, which is a known challenge in syntactic parsing, will lead to different maximum boundaries for a single mention.

In order to decouple coreference evaluation from maximum boundary detection complexities,
smaller corpora like MUC \cite{hirschman97}, ACE \cite{ace2002data}, and ARRAU \cite{arrau}
explicitly annotate the minimum span as well as the maximum logical span of each mention.
The annotated minimum spans indicate the minimum strings that a coreference resolver must identify for the corresponding mentions. 
This solution comes with an additional annotation cost.
As a result, the annotation of minimum spans has been discarded in larger corpora like CoNLL-2012 \cite{2012-ontonotes}.

In this paper, we propose \mina, a MINimum span extraction Algorithm that
automatically determines minimum spans from constituency-based parse trees.
Based on our analyses,
\mina spans are compatible with those that are manually annotated by experts.
By using \mina, we can benefit from minimum span evaluation
for all corpora without introducing additional annotation costs.

While the use of \mina spans already benefits in-domain evaluation, by reducing the gap between the performance on gold vs.\ system mentions, it has a more significant impact on cross-dataset evaluation, in which detected maximum mention boundaries are noisier due to domain shift.

Cross-dataset coreference evaluation is used to assess the generalization of coreference resolvers \cite{moosavi-strube-2017-lexical,moosavi-strube-2018-using}. 
Coreference resolution is a mid-step for text understanding in downstream tasks, e.g., question answering, text summarization, and information retrieval. 
Therefore, generalization is an important property for coreference resolvers because downstream datasets are not necessarily from the same domain as those of coreference-annotated corpora. 

When coreference resolvers are applied to a new domain, detected maximum boundaries become noisier, e.g., gold and system mentions differ by the inclusion or exclusion of surrounding commas or quotation marks.
Such noisy boundaries directly affect the coreference evaluation scores based on maximum spans.
The use of minimum spans reduces the impact of such noises in coreference evaluation and results in more reliable comparisons between different coreference resolvers.





\section{Boundary Mismatch Example}
\begin{figure}[!htb]
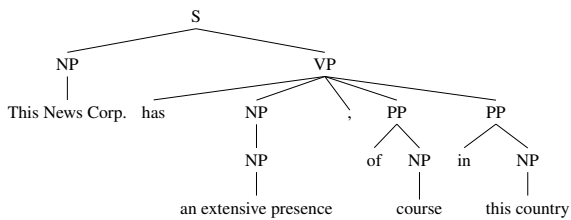

\centering
 \resizebox{0.98\linewidth}{!}{\Tree [.S [.NP \text{This News Corp.} ] [.VP has [.NP [.NP \text{an extensive presence} ]] , [.PP of [.NP course ]][.PP in [.NP \text{this country} ]]]]}
\caption{Gold parse tree of Example~\ref{example}.\label{extensive_presence_gold_parse}}
\end{figure}

\begin{figure}[!htb]
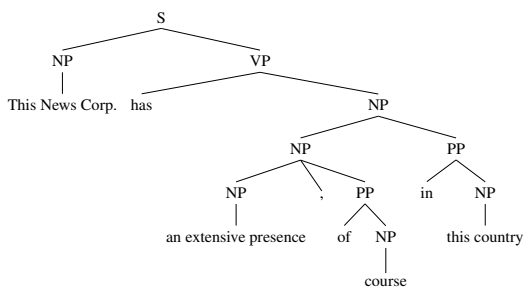

\centering
 \resizebox{0.9\linewidth}{!}{\Tree [.S [.NP \text{This News Corp.} ][.VP has [.NP [.NP [.NP \text{an extensive presence} ] , [.PP of [.NP course ]]] [.PP in [.NP \text{this country} ]]]]]}
\caption{System parse tree of  Example~\ref{example}.\label{extensive_presence_sys_parse}}
\end{figure}

Example~\ref{example}, and its corresponding gold and system parse trees in Figure~\ref{extensive_presence_gold_parse} and Figure~\ref{extensive_presence_sys_parse}, respectively, show a sample boundary mismatch from the CoNLL-2012 development set.
Based on the gold parse tree (Figure~\ref{extensive_presence_gold_parse}),
``an extensive presence'' is the maximum span of the first coreferring mention in Example~\ref{example}.
However, the corresponding maximum boundary for this same mention is ``an extensive presence, of course in this country'' based on the system parse tree (Figure~\ref{extensive_presence_sys_parse}).

\begin{example}
 This News Corp. has [an extensive presence]$_{1}$, of course in this country.
 [That presence]$_{(1)}$ may be expanding soon.
 \label{example}
\end{example}


A system that uses the system parse tree for mention detection
links ``that presence'' to ``an extensive presence, of course in this country''
and gets penalized based on recall and precision. 
This penalty is the same as that of a system that links ``that presence'' to ``this News Corp.''.
Recall drops because of not recognizing ``an extensive presence''
and precision drops because of detecting a spurious mention.

\section{Background}
\label{sect:rel}
\mina is an attempt to decouple coreference evaluation from parsing errors
to some extent.
This motivation is the same as the one that resulted in the manual annotation of minimum spans in the 
MUC, ACE and ARRAU corpora.
According to the MUC task definition,\footnote{\url{http://www-nlpir.nist.gov/related_projects/muc/proceedings/co_task.html}}
the use of minimum spans in coreference evaluation is as follows:\footnote{The ARRAU dataset also follows this way of using minimum spans.}
Assume $m_{max}$ and $m_{min}$ are the annotated maximum and minimum spans for the mention $m$.
The system mention $\hat{m}$ is equivalent to $m$ if it includes $m_{min}$ and it does not 
include any tokens beyond those that are included in $m_{max}$.
This way of using minimum spans does not handle inconsistencies in gold vs. system mention boundaries 
in which system boundaries are larger than their corresponding gold boundaries, 
as it is the case for the mention ``an extensive presence, of course in this country'' in Example~\ref{example}.

Compared to manually annotated minimum spans:
\squishlist
\item \mina is applicable to any English coreference corpus.\footnote{We did not have the manually annotated minimum spans for coreference corpora of other languages in order to verify whether \mina is also applicable to them.} In contrast, manually annotated minimum spans can be only used in their own corpora.
\item For coreference evaluation, \mina extracts minimum spans for both gold and system mentions based on a single parse tree. Therefore, it can handle system-detected maximum spans that are either shorter or longer than their corresponding gold maximum span.  
\squishend

The coreference resolver of \newcite{penghaoruo15} is developed around the idea
that working with mention heads is more robust compared to working with maximum mention boundaries.
In this regard, they develop a system that resolves coreference relations based on mention heads.
The resolved mention heads are then expanded to full mention boundaries
using a separate classifier that is trained to do so. 
\newcite{penghaoruo15} also report the evaluation scores 
using both maximum mention boundaries and mention heads.
\newcite{penghaoruo15} extract mention heads using Collins' head finder rules \cite{collins99b}.
They use gold constituency-based parse trees and gold named entity information.
The gold parse information is only used during training to train their mention head detection classifier.
The gold named entity information is used to specify the whole span of named entities as their heads.
The reason is that the head finding rules only specify one word as a head,
and one-word heads can be troublesome for named entities,
e.g.,\ ``Green'' would be selected as the head of both ``Mary Green'' and ``John Green''.

In this paper, we also examine the use of head words as minimum spans.
We show that compared to head words, \mina spans are more compatible with expert annotated minimum spans.

Since we evaluate minimum spans on various corpora, from which some do not include gold named entity information or even gold parse trees,
we only use Collins' head finder rules, without the final adjustment for named entities, as the baseline for minimum span detection.

Collins' rules for finding the head of a noun phrase (NP) are as follows:
\squishlist
\item If the tag of the last word is POS, return it as the head,
\item else return the first child, from right to left, with an NN, NNP, NNPS, NNS, NX, POS, or JJR tag, if there is any,
\item else return the first child, from left to right, with an NP tag, if there is any,
\item else return the first child, from right to left, with one of the \$, ADJP, or PRN tags, if there is any,
\item else return the first child, from right to left, with a CD tag, if there is any,
\item else return the first child, from right to left, with a JJ, JJS, RB, or QP tag, if there is any,
\item else return the last word.
\squishend

For the head finder rules for phrases other than NPs, please refer to Appendix A of \newcite{collins99b}.

\section{How to Determine Minimum Spans?}
\label{sect:how}
We process the constituency-based parse trees of mentions, i.e.,\ the parse sub-tree of their corresponding maximum span, in a breadth-first manner to determine minimum spans.
Algorithm~\ref{min-span-algo} outlines the minimum span extraction process.
In this algorithm, \texttt{root} is the root of the mention's parse tree, \texttt{tags} is the set of acceptable syntactic tags for extracting minimum spans, \texttt{min-depth} is the depth of the minimum span nodes in the parse tree, and \texttt{min-spans} is the output of the algorithm that corresponds to the set of mention words that belong to the minimum span.
\texttt{min-depth} is initially set to $\infty$, and \texttt{tags} and \texttt{min-spans} are empty.

\begin{algorithm}[!htb]
\footnotesize
\SetAlgoLined
\DontPrintSemicolon 
\SetKwFunction{algo}{MINA}
\SetKwProg{myalg}{Algorithm}{}{}
  \myalg{\algo{root}}{
  
   min-depth=$\infty$\;
     
   \If{tags=$\emptyset$}{
       \If{\text{root is an NP}}{
             \text{tags= \{NP acceptable tags\}}}
          \ElseIf{\text{root is a VP}} {
             \text{tags=\{VP acceptable tags\}}\;
          }
       }
   \text{Process $root$ in a breadth-first manner} \;
   \For{each processed node n \text{ } }{
   \If{n.tag $\not\in$ tags \text{ }}{
       \text{skip processing n's children} \;
   }
   \ElseIf{n \text{is an acceptable terminal node and} n.depth $\leq$ min-depth \text{ }}{
      min-spans.add(n)\;
	  min-depth=n.depth \;
    }
   }

  }
  
\caption{Extraction of minimum spans.}
\label{min-span-algo}
\end{algorithm}

The set of acceptable terminal nodes in a parse tree are those that include at least one word other than a determiner\footnote{A word with the ``DT'' POS tag.} or a conjunction\footnote{A word with the ``CC'' POS tag.}.
We do not further split terminal nodes, e.g., an acceptable terminal node may contain both a determiner as well as a noun.
For extracting the minimum span of a noun phrase, the set of acceptable syntactic tags is \{``NP'' (noun phrase), ``NML'' (nominal modifier), ``QP'' (quantifier phrase used within NP), ``NX'' (used within certain complex NPs)\}. For verb phrases, ``VP'' is the only acceptable tag.

\mina processes the parse tree in a breadth-first manner.
It skips processing sub-trees that are rooted by a node whose syntactic tag is not acceptable, e.g., ``PP''.
For the rest of the nodes, it extracts all acceptable terminal nodes that have the shortest distance to \texttt{root} as minimum spans.



\begin{figure}[!htb]
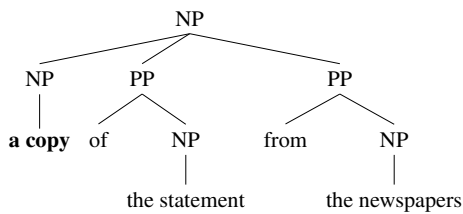

\centering
 \resizebox{0.8\columnwidth}{!}{\Tree [.NP [.NP \textbf{a copy} ][.PP of [.NP \text{the statement} ]][.PP from [.NP \text{the newspapers} ]]]}
\caption{\mina span in an NP with the grammar form ``NP $->$ NP PP PP''. \mina span is boldfaced.\label{fig:mina-example5}}
\end{figure}

For instance, in Figure~\ref{fig:mina-example5}, the root node is an NP and \texttt{tag} would be set to NP's acceptable tags. Therefore, among the children of the root, \mina would only process the child with an NP tag (the left child) and skip the one with the PP tag.  

If the final minimum span is empty, e.g., if due to parsing errors the syntactic tag of none of the tree nodes is among the acceptable tags, we fall back to using the maximum span.\footnote{If we use gold parse trees, 
this happens for 14 mentions in the CoNLL-2012 development set
from which ten are one-word mentions,
e.g., ``ours'' is detected as ``ADJP''.}

\begin{figure}[!htb]
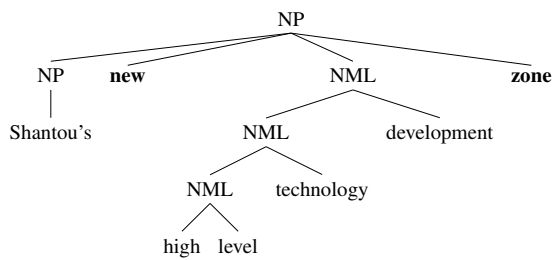

\centering
 \resizebox{0.95\columnwidth}{!}{\Tree [.NP [.NP Shantou's ] \textbf{new} [.NML [.NML [.NML high level ] technology ] development ] \textbf{zone} ]}
\caption{\mina spans in an NP with a nested structure.\label{fig:mina-example2}}
\end{figure}
\begin{figure}[!htb]
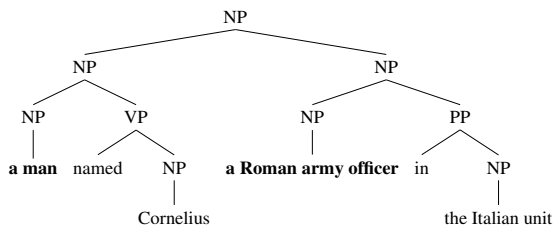

\centering
 \resizebox{0.95\columnwidth}{!}{\Tree [.NP [.NP [.NP \textbf{a man} ][.VP named [.NP Cornelius ]]][.NP [.NP \textbf{a Roman army officer} ][.PP in [.NP \text{the Italian unit} ]]]]}
\caption{\mina spans in an appositive noun phrase.\label{fig:mina-example3}}
\end{figure}
\begin{figure}[!htb]
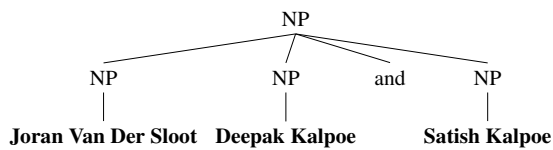

\centering
 \resizebox{0.95\columnwidth}{!}{\Tree [.NP [.NP \textbf{Joran Van Der Sloot} ][.NP \textbf{Deepak Kalpoe} ] and [.NP \textbf{Satish Kalpoe} ]]}
\caption{\mina spans in an NP with conjunction. Boldfaced minimum spans belong to a single mention.\label{fig:mina-example4}}
\end{figure}

\paragraph{MINA extraction examples.}
Figures~\ref{fig:mina-example5}-\ref{fig:mina-example4} show the \mina minimum spans of various noun phrases with different internal structures. The corresponding \mina spans of the parse trees are boldfaced. 


\paragraph{Using MINA for coreference evaluation.} 
For each coreference evaluation, we have a key file, including gold coreference annotations, and a system file, including predicted coreference outputs.
For coreference evaluation using minimum spans, we use the provided parse trees in the key file.\footnote{If the key file does not include parse information, we parse it with the Stanford parser.} 
Therefore, the minimum spans of both gold mentions and system mentions are determined based on the same parse tree.
We then use minimum spans instead of maximum spans in all scoring metrics, i.e.,\ a gold and a system mention are considered equivalent if they have the same minimum span.

\begin{figure}[!htb]
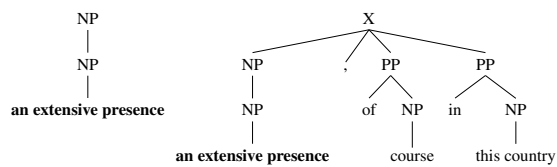

\centering
 \resizebox{0.95\columnwidth}{!}{\Tree [.NP [.NP \textbf{an extensive presence} ]] \Tree[.X [.NP [.NP \textbf{an extensive presence} ]] , [.PP of [.NP course ]][.PP in [.NP \text{this country} ]]]}
\caption{Mention trees of Example~\ref{example}. The left and right sub-trees represent the boundaries of gold and system mentions, respectively. \label{fig:example-1-min-spans}}
\end{figure}

The corresponding sub-trees of the discussed gold and system mentions of Example~\ref{example}, based on the gold parse tree in Figure~\ref{extensive_presence_gold_parse}, are shown in Figure~\ref{fig:example-1-min-spans}.\footnote{If the boundary of a mention is not recognized as a single phrase in the parse tree, as it is the case for the system mention, we add a dummy root (``X'' in the right subtree of Figure~\ref{fig:example-1-min-spans}) to include the whole span into a single phrase.}
The \mina span of both of these two trees is ``an extensive presence''. Therefore, the gold coreference chain \{``an extensive presence'', ``that presence''\} and the system coreference chain \{``an extensive presence, of course in this country'', ``that presence''\} are equivalent if they are evaluated based on minimum spans.

\section{Evaluating MINA Spans}
In order to analyze the detected \mina spans, we evaluate the following two properties: 
\squishlist
\item Length of \mina spans: since we retrieve \mina spans from the corresponding parse tree of the mentions, \mina spans are always a subset of their corresponding maximum span words. 
However, on average, the length of minimum spans (number of containing words) should be smaller than that of maximum spans.
\item Compatibility of automatically extracted minimum spans with those that are manually annotated by experts: we evaluate \mina spans against manually annotated minimum spans, called \texttt{MIN}, in the MUC and ARRAU corpora to examine whether the reduced spans still contain words of the mention that were deemed important by experts. 
\squishend

For the experiments of this section, we use the MUC-6, MUC-7, ARRAU, and CoNLL-2012 corpora, from which MUC and ARRAU contain manually annotated minimum spans.
We use the Stanford neural constituency parser \cite{socher13parse} for getting system parse trees, unless otherwise stated.
For the ARRAU corpus, we use mentions of the training split of the RST Discourse Treebank subpart.

As a baseline, we also evaluate the syntactic head of mentions, based on Collins' rules, as the minimum span.\footnote{We use the implementation of the head-finding rules that is available at \url{https://github.com/smartschat/cort/}.}


\paragraph{How does the length of evaluated spans change by using MINA?}
Table~\ref{tab:min-span-head-len} shows the average length of maximum spans vs. that of \mina spans on the training splits of the MUC-6, MUC-7 and ARRAU corpora as well as the development set of the CoNLL-2012 dataset. 
For the CoNLL-2012 dataset, we use the provided gold parse information.
We parse the MUC and ARRAU datasets,
since the gold parse information is not available for these datasets.

Based on Collins' head finder rules, the detected head always includes one word.

\begin{table}[!htb]
    \begin{center}
        \resizebox{\columnwidth}{!}{%

    \begin{tabular}{l|llll}
		& MUC-6 & MUC-7 & ARRAU & CoNLL\\ \hline
		maximum span & $5.2$ & $5.3$ & $3.8$ & $2.4$\\ \hline
		\mina span & $2.6$ & $2.7$ & $2.0$ & $1.6$\\ 
    \end{tabular}
}
    \end{center}
    \caption{The average length of \mina spans compared to that of maximum spans in the MUC, ARRAU, and CoNLL-2012 datasets.}
    \label{tab:min-span-head-len}
\end{table}

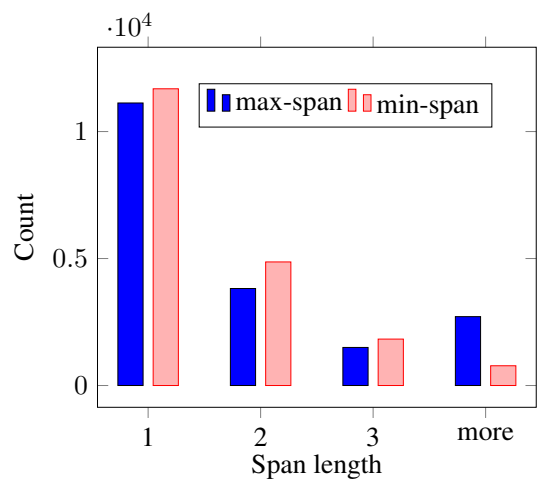
\begin{figure}
    \centering
\resizebox{0.92\columnwidth}{!}{%
\begin{tikzpicture}
\begin{axis}[
    ybar,
    enlargelimits=0.15, 
    legend style={at={(0.9,0.9)},
      legend columns=-1},
    xlabel={Span length},   ylabel={Count},  ylabel near ticks,
    symbolic x coords={1,2,3,more},
    xtick=data, ybar=4pt, width=\columnwidth,
    nodes near coords align={vertical},
    ]
\addplot [fill=blue] coordinates {(1,11125) (2,3820) (3,1499) (more,2712)};
\addplot coordinates {(1,11684) (2,4866) (3,1828) (more,778)};
\legend{max-span,min-span}
\end{axis}
\end{tikzpicture}
}
\caption{Span length based on maximum vs. \mina's minimum spans on the CoNLL-2012 development set.}
\label{fig:span_length}
\end{figure}

Figure~\ref{fig:span_length} shows the number of mentions with the length of one, two, three, and $\geq$4 based on both maximum and \mina spans on the CoNLL-2012 development set. 
The length of the maximum span of around $14\%$ of mentions is longer than three, while this ratio is only $4\%$ for \mina minimum spans.  
Mentions with long \mina spans include appositions or conjunctions, e.g.,\ the \mina span in Figure~\ref{fig:mina-example4}.

\paragraph{Does MINA correlate with MIN?}
We evaluate \mina minimum spans against manually annotated minimum spans in the MUC and ARRAU corpora.
The manually annotated minimum span in these corpora is referred to as \texttt{MIN}.

Table~\ref{tab:min-span-head-min} shows the ratio of minimum spans that contain the corresponding \texttt{MIN} when the minimum span is extracted by \mina and the head finding rules. 
As we can see, \mina contains \texttt{MIN} in the majority of the mentions, and therefore, it is compatible with what experts would consider as the most important part of the mentions.

\begin{table}[!htb]
    \begin{center}
    \begin{tabular}{l|lll}
		& MUC-6 & MUC-7 & ARRAU\\ \hline
		\mina & $96.2$ & $93.1$ & $98.3$ \\ \hline
		head & $94.0$ & $91.1$ & $93.9$ \\
    \end{tabular}

    \end{center}
    \caption{Ratio of detected \mina and head words which contain the corresponding \texttt{MIN} annotations in the MUC and ARRAU corpora. The same parse information is used for detecting both \mina and head words. 
    Datasets are parsed using the Stanford neural parser.
    }
    \label{tab:min-span-head-min}
\end{table}

\begin{figure}[!htb]
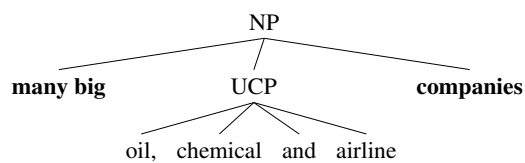

\centering
 \resizebox{0.9\columnwidth}{!}{\Tree[.NP \textbf{many big} [.UCP oil, chemical and airline ] \textbf{companies} ]}
\caption{System parse tree of a mention from ARRAU. \mina spans are boldfaced. ``many'' and ``companies'' are the corresponding \texttt{MIN} and head, respectively. \label{fig:arrau-mina-vs-head}}
\end{figure}

Figure~\ref{fig:arrau-mina-vs-head} shows an example from ARRAU in which \mina contains \texttt{MIN} but the head does not.

\begin{figure}[!htb]
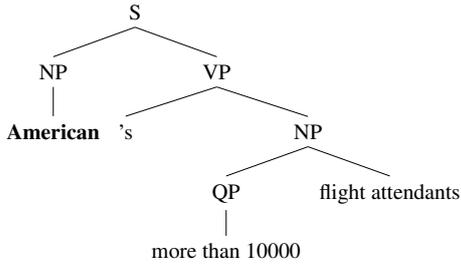

\centering
 \resizebox{0.8\columnwidth}{!}{\Tree [.S [.NP \textbf{American} ][.VP 's [.NP [.QP \text{more than 10000} ] \text{flight attendants} ]]]}
\caption{The system parse tree of a mention from MUC-6. \mina spans are boldfaced (``American''). ``attendants'' is the annotated MIN.\label{fig:muc6-parsing-error}}
\end{figure}

\begin{figure}[!htb]
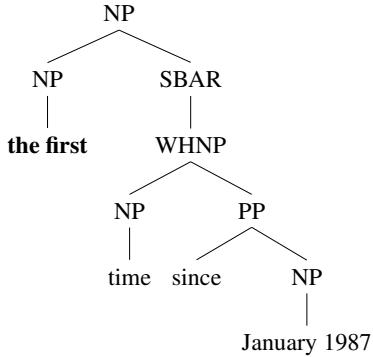

\centering
 \resizebox{0.65\columnwidth}{!}{\Tree [.NP [.NP \textbf{the first} ][.SBAR [.WHNP [.NP time ][.PP since [.NP \text{January 1987} ]]]]]}
\caption{The system parse tree of a mention from ARRAU. \mina spans are boldfaced (``the first''). ``time'' is the annotated MIN.\label{fig:arrau-parsing-error}}
\end{figure}

\mina and \texttt{MIN} inconsistencies, i.e.,\ cases in which \mina does not contain \texttt{MIN}, are mainly due to parsing errors. 
Figure~\ref{fig:muc6-parsing-error} and Figure~\ref{fig:arrau-parsing-error} show two examples from the MUC and ARRAU datasets in which \mina selects an incorrect minimum span because of an incorrect parse tree.  



\begin{figure}[!htb]
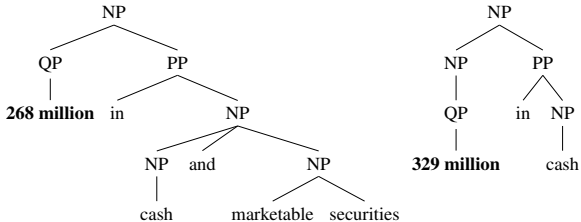

\centering
 \resizebox{\columnwidth}{!}{\Tree [.NP [.QP \textbf{268 million} ][.PP in [.NP [.NP cash ] and [.NP marketable securities ]]]] \Tree[.NP [.NP [.QP \textbf{329 million} ]][.PP in [.NP cash ]]]}
\caption{The system parse trees of two mentions from ARRAU. \mina spans are boldfaced. ``securities'' and ``cash'' are annotated as \texttt{MIN} for the left and right mentions, respectively.  \label{fig:arrau-true-mismatch}}
\end{figure}

Figure~\ref{fig:arrau-true-mismatch} shows two sample mismatch examples between \mina and \texttt{MIN} from the ARRAU dataset, in which the mismatch is not due to parsing errors.

In order to investigate the effect of using a different parser,
we perform the experiment of Table~\ref{tab:min-span-head-min} using the Stanford English PCFG parser \cite{klein03b}.
The results are reported in Table~\ref{tab:min-span-head-min-pcfg}.
As we see, the use of a better parser, i.e.,\ the Stanford neural parser, makes \mina spans, as well as detected heads, more consistent compared to \texttt{MIN} spans.

\begin{table}[!htb]
    \begin{center}
    \begin{tabular}{l|rrr}
		& MUC-6 & MUC-7 & ARRAU\\ \hline
		\mina & $95.6$ & $92.4$ & $98.1$ \\ \hline
		head & $92.9$ & $90.0$  & $93.4$ \\
    \end{tabular}

    \end{center}
    \caption{Ratio of the detected \mina and head words that contain their corresponding \texttt{MIN} annotations in MUC and ARRAU. \mina and head words are detected using the parse trees of the Stanford PCFG parser.}
    \label{tab:min-span-head-min-pcfg}
\end{table}





In addition to the above two properties, i.e.\ the length of minimum spans and their consistency with \texttt{MIN} annotations, we also check that \mina returns \emph{different minimum spans for distinct overlapping mentions}.

As an example, the minimum span of the mention ``John and Mary'' should be different from those of ``John'' and ``Mary'', because they all refer to different entities. In this regard, we examine all overlapping coreferent mentions in the CoNLL-2012 English development set, from which none of the overlapping mentions has the same \mina span. However, this is not the case for heads.

\begin{table*}[htbp]
    \begin{center}
	   \begin{tabular}{l|lll||lll}
	  \multicolumn{1}{c}{} & \multicolumn{3}{c}{CoNLL} & \multicolumn{3}{c}{LEA}\\  
	  \multicolumn{1}{c}{} & max & \mina & head & max & \mina & head \\
     \hline
     \multicolumn{1}{c}{} &  \multicolumn{6}{c}{CoNLL-2012 test set} \\ \hline
    Stanford rule-based & $55.60$ (8) & $57.55$ (8)& $57.38$ (8) & $47.31$ (8) & $49.65$ (8) & $49.44$ (8)\\
     cort & \cellcolor{blue!10}$63.03$  (7)& \cellcolor{blue!10}$64.60$ (6)& $64.51$ (6) & $56.10$  (6)& $58.05$ (6)& $57.93$ (6)\\ 
    Peng et al. & \cellcolor{blue!10}$63.05$ (6)& \cellcolor{blue!10}$63.50$ (7)& $63.54$ (7)& $55.22$ (7) & $55.76$ (7) & $55.80$ (7)\\ 
     deep-coref ranking & $65.59$ (5) & $67.29$ (5)& $67.09$ (5)& $59.58$ (5) & $61.70$ (5) & $61.43$ (5)\\
     deep-coref RL & $65.81$ (4) & $67.50$ (4)& $67.36$ (4)& $59.76$ (4) & $61.84$  (4)& $61.64$ (4)\\
     Lee et al. 2017 single & $67.23$ (3) & $68.55$ (3) & $68.53$ (3) & $61.24$ (3)& $62.87$  (3)& $62.82$ (3)\\
     Lee et al. 2017 ensemble & $68.87$ (2)  & $70.12$ (2) & $70.05$ (2) & $63.19$ (2) & $64.76$ (2)& $64.64$ (2)\\
	 Lee et al. 2018 & $72.96$ (1) & $74.26$ (1)& $75.29$ (1)& $67.73$ (1) & $69.32$ (1)& $70.40$ (1)\\
     \hline
     \multicolumn{1}{c}{} &  \multicolumn{6}{c}{WikiCoref} \\ \hline
     Stanford rule-based & \cellcolor{blue!10}$51.78$ (4)& \cellcolor{blue!10}$53.79$ (5)& $57.10$ (4) & \cellcolor{blue!10}$43.28$ (5)  & \cellcolor{blue!10}$45.48$ (6) & $49.28$ (4)\\
     deep-coref ranking & \cellcolor{blue!10}$52.90$ (3) & \cellcolor{blue!10}$55.16$ (2) &$57.13$ (3)& $44.40$ (3)  & $46.98$ (3) & $49.05$ (5)\\
     deep-coref RL & \cellcolor{blue!10}$50.73$ (5) & \cellcolor{blue!10}$54.26$ (4)& $57.16$ (2)& \cellcolor{blue!10}$41.98$ (6)  & \cellcolor{blue!10}$46.02$ (4)& $49.29$ (3)\\
	 Lee et al. 2017 single & $50.38$ (6) & $52.16$ (6)& $54.02$ (6) & \cellcolor{blue!10}$43.86$ (4) & \cellcolor{blue!10}$45.75$ (5) & $47.69$ (6)\\
	 Lee et al. 2017 ensemble & \cellcolor{blue!10}$53.63$ (2) & \cellcolor{blue!10}$55.03$ (3) & $56.80$ (5) & $47.50$ (2)  & $48.98$ (2) & $50.87$ (2)\\
	 Lee et al. 2018 & $57.89$ (1) & $59.90$ (1) & $61.33$ (1)& $52.42$ (1) & $54.63$ (1) & $56.19$ (1)\\
    \end{tabular}
    \end{center}
    \caption{Evaluations based on maximum span, \mina, and head spans on the CoNLL-2012 test set and WikiCoref. The ranking of corresponding scores is specified in parentheses. Rankings which are different based on maximum vs. \mina spans are highlighted.}
    \label{tab:state_of_the_art}
\end{table*}

\section{Effect on Coreference Evaluation}
\label{ch:mint:analysis}
\subsection{Experimental Setup}
In this section, we investigate how the use of minimum spans instead of maximum spans in coreference evaluation affects the results in in-domain as well as cross-dataset evaluations.
For comparisons, we use the 
\emph{CoNLL} score \cite{pradhan14},
i.e.\ the average F$_1$ value of \emph{MUC} \cite{vilain95}, \emph{B}$^3$ \cite{bagga98b},
and \emph{CEAF}$_e$ \cite{luoxiaoqiang05a}, and the \emph{LEA} F$_1$ \cite{moosavi16b} score.\footnote{We use the python implementation that is available at \url{https://github.com/ns-moosavi/coval}.}  
Minimum spans are detected using both \mina and Collins' head finding rules.
All examined coreference resolvers are trained on the CoNLL-2012 training data.
For in-domain evaluations, models are evaluated on the CoNLL-2012 test data and minimum spans are extracted using gold parse trees, which are provided in CoNLL-2012.\footnote{We also examined the in-domain results of Table~\ref{tab:state_of_the_art} based on the system parse trees of CoNLL-2012 instead of gold parse trees. The differences between scores based on \mina spans that are extracted from gold vs.\ those that are extracted from system parse trees were only about 0.2 points.}

For cross-dataset evaluations, models are tested on the WikiCoref dataset \cite{ghaddar16b}.
For extracting minimum spans, we parse WikiCoref by the Stanford neural parser.
This dataset is annotated using the same annotation guidelines as that of CoNLL-2012, however, it contains documents from a different domain. 
CoNLL-2012 contains the newswire, broadcast news, broadcast conversation, telephone conversation, magazine, weblogs, and Bible genres while the annotated documents in WikiCoref are selected from Wikipedia.


%

\subsection{Results}
Table~\ref{tab:state_of_the_art} shows the maximum vs.\ minimum span evaluations of several recent coreference resolvers on the CoNLL-2012 test set and the WikiCoref dataset.
The examined coreference resolvers are as follows: the Stanford rule-based system \cite{leeheeyoung13},
the coreference resolver of \newcite{penghaoruo15},
the ranking model of \texttt{cort} \cite{martschat15c}, the ranking and reinforcement learning models of \texttt{deep-coref} \cite{clarkkevin16a,clarkkevin16b}, 
the single and ensemble models of \newcite{leekenton17}, and the current state-of-the-art system by \newcite{N18-2108}.

We make the following observations based on the results of Table~\ref{tab:state_of_the_art}:

\paragraph{Using minimum spans in coreference evaluation strongly affects the comparisons in the cross-dataset setting.}
The results on the WikiCoref dataset show that mention boundary detection errors specifically affect coreference scores in cross-dataset evaluations.
The ranking of systems is very different by using maximum vs. minimum spans.
The reinforcement learning model of \texttt{deep-coref}, i.e.,\ \texttt{deep-coref RL}, has the most significant difference when it is evaluated based on maximum vs.\ minimum spans (about 4 points).
The ensemble model of \texttt{e2e-coref}, on the other hand, has the least difference between maximum and minimum span scores (1.4 points), which indicates it better recognizes maximum span boundaries in out-of-domain data.

\paragraph{Using minimum spans in coreference evaluation reduces the gap between the performance on gold vs.\ system mentions.}
It is shown that there is a large gap between the performance of a coreference resolver on gold vs.\ system mentions, see e.g., \newcite{penghaoruo15}.
The use of minimum spans in coreference evaluation reduces this gap by about two points. The comparison of the results of different systems on gold and system mentions using both maximum and minimum spans are included in Appendix~\ref{sec:appendix}.

\paragraph{Evaluation based on minimum spans reduces the differences that are merely due to better maximum boundary detection.}
The coreference resolver of \newcite{penghaoruo15} 
has the smallest difference between its maximum and minimum span evaluation scores.
This result indicates the superiority of \newcite{penghaoruo15}'s mention boundary detection method compared to other approaches.\footnote{It has a separate classifier for detecting maximum boundaries based on mention heads.}
Based on maximum spans, \newcite{penghaoruo15} performs on-par with \texttt{cort} while \texttt{cort} outperforms it by about one percent when they are evaluated based on minimum spans.
Therefore, the use of minimum spans in coreference evaluation decreases the effect of mention boundary detection errors in coreference evaluation and results in fairer comparisons.

\section{Analysis}
In order to better understand the impact of using minimum spans in cross-dataset evaluations, we analyze the output of \texttt{deep-coref RL}, on which minimum span evaluation has the largest impact, for the cases in which a system mention and its corresponding gold mention have the same minimum span while they have different maximum boundaries.

We have included some examples from these mismatches in Example~\ref{appositive}--Example~\ref{last}.
The boundaries of gold and system mentions are determined by $g$ and $s$ indices, respectively. Mismatching spans are boldfaced in all examples.

We observe that the majority of the mismatches are due to (1) incorrect detection of appositive relation (Example~\ref{appositive}), (2) mismatch as a result of not including a surrounding quotation (Example~\ref{quotation}), and (3) inclusion of an additional comma at the end of the mention (Example~\ref{comma}).

\begin{example}
 Canada is noted for having a positive relationship with [[the Netherlands]$_g$\textbf{, owing,}]$_s$ in part, to its contribution to the Dutch liberation during World War II.
 \label{appositive}
\end{example}

\begin{example}
 .[[Le Courrier du Sud]$_g$\textbf{,}]$_s$ published by Quebecor Media, is the oldest, and contains inserts tailored to specific boroughs
 \label{comma}
\end{example}

\begin{example}
in 2007, [[Pierce College]$_g$ \textbf{sheltered}]$_s$ and fed more than 150 horses under the direction of the L.A. County Volunteer Equine Response team.
\end{example}

\begin{example}
Prime Minister Brian Mulroney's Progressive Conservatives abolished the NEP and changed the name of FIRA to [\textbf{``}[Investment Canada]$_s$\textbf{''}]$_g$, to encourage foreign investment.
\label{quotation}
\end{example}

\begin{example}
In 2011, [$_s$\textbf{nearly 6.8 million} [$_g$Canadians]] listed a non-official language as their mother tongue.
\label{last}
\end{example}


\section{Conclusions}
Coreference evaluation based on maximum spans directly penalizes coreference resolvers because of parsing complexities and also small noises in mention boundary detection.
This is a known problem that is addressed by manually annotating minimum spans as well as maximum spans in several corpora.
Minimum span annotation is expensive, and therefore 
it is not a scalable solution for large coreference corpora.
In this paper, we propose the \mina algorithm to automatically extract minimum spans
without introducing additional annotation costs.
\mina automatically extracts corresponding minimum spans for both gold and system mentions and uses the resulting minimum spans in the standard evaluation metrics.
Based on our analysis on the MUC and ARRAU datasets, 
extracted minimum spans are compatible
with those that are manually annotated by experts.
The incorporation of automatically extracted minimum spans reduces the effect of maximum boundary detection errors in coreference evaluation and results in a fairer comparison.
Our results show that the use of minimum spans in coreference evaluation is of particular importance for cross-dataset settings, in which the detected maximum boundaries are noisier.

In addition to coreference evaluation, automatically extracted minimum spans can benefit 
the annotation process of new corpora.
If we provide automatically extracted minimum spans alongside maximum spans to the annotators,
the annotation of coreference relations may get easier.
For instance, detecting the coreference relation of the two nested mentions in ``[a deutsche mark based currency board where we have a foreign governor on [the board]$_{(1)}$]$_{(1)}$''\footnote{Taken from the CoNLL-2012 development set.}
would be more straightforward knowing that the minimum span of the first mention is ``a currency board''.

A future direction is to investigate the effect of using \mina spans not only in evaluation but also for training existing coreference resolvers. Maximum spans are recoverable given the \mina spans and their corresponding parse trees. Therefore, we can use \mina spans for training and testing coreference models and then retrieve their corresponding maximum spans for evaluation.
Investigating the use of \mina in other NLP areas, e.g., evaluating spans in named entity recognition or reading comprehension, is another future line of work.

\section*{Acknowledgments} The authors would like to thank Mark-Christoph M\"uller, Ilia Kuznetsov and the anonymous reviewers for their valuable comments and feedbacks. 
This work has been supported by the Klaus Tschira Foundation, Heidelberg,
Germany, the German Research Foundation (DFG) as part of the QA-EduInf project (grant GU 798/18-1 and grant RI 803/12-1), and  the DFG-funded research training group “Adaptive Preparation of Information form Heterogeneous Sources” (AIPHES, GRK 1994/1). 
\bibliography{mybib}
\bibliographystyle{acl_natbib}

\appendix

\section{Appendix}
\label{sec:appendix}

\begin{table*}[!htb]
    \begin{center}
	   \begin{tabular}{l|lll||lll}
	  \multicolumn{1}{c}{} & \multicolumn{3}{c}{CoNLL} & \multicolumn{3}{c}{LEA}\\  
	  \multicolumn{1}{c}{} & max & \mina & head & max & \mina & head \\
    \hline
    fernandes &$60.6$ (1) & $62.2$ (1) & $63.9$ & $53.3$  & $55.1$   & $57.0$ \\ 
    martschat & $57.7$ (2) & $59.7$ (2)  & $61.0$ & $50.0$ & $52.4$  & $53.9$ \\ 
    bjorkelund&  $57.4$ (3) & $58.9$ (3) & $60.7$ & $50.0$ & $51.6$  & $53.6$ \\ 
    chang & $56.1$ (4) & $58.0$ (4) & $59.6$ & $48.5$ & $50.7$  & $52.5$ \\ 
    chen& $54.5$ (5) & $56.5$ (5) & $58.2$ & \cellcolor{blue!10}$46.2$ & \cellcolor{blue!10}$48.6$   & $50.4$ \\ 
    chunyuang& $54.2$ (6) & $56.1$ (6) & $57.9$ & $45.8$ & $48.1$   & $50.2$ \\ 
    shou & \cellcolor{blue!10}$53.0$ (7) & \cellcolor{blue!10}$54.8$ (8) & $56.5$  & $44.0$ & $46.1$  & $48.1$ \\ 
    yuan & \cellcolor{blue!10}$52.9$ (8) & \cellcolor{blue!10}$54.9$ (7) & $56.7$ & $44.8$ & $47.0$  & $48.9$ \\
    xu &$52.6$ (9) & $53.9$ (9) & $55.2$ & \cellcolor{blue!10}$46.8$  & \cellcolor{blue!10}$48.4$  & $50.0$ \\ 
    uryupina &\cellcolor{blue!10}$50.0$ (10) & \cellcolor{blue!10}$51.0$ (11) & $52.4$ & $41.2$  & $42.3$  & $43.7$ \\ 
    songyang & \cellcolor{blue!10}$49.4$ (11)  & \cellcolor{blue!10}$51.3$ (10) & $52.9$ & $41.3$ & $43.5$  & $45.3$ \\ 
    \end{tabular}
    \end{center}
    \caption{CoNLL-2012 shared task systems evaluations based on maximum spans, \mina spans, and head words. The rankings based on the CoNLL scores are included in parentheses for maximum and \mina spans. Rankings which are different based on maximum vs. \mina spans are highlighted. }
    \label{conll-2012-min-max-evaluations}
\end{table*}

Table~\ref{conll-2012-min-max-evaluations} shows 
\emph{CoNLL} scores and the \emph{LEA} F$_1$ values 
of the participating systems in the CoNLL-2012 shared task (closed task with predicted syntax and mentions) based on both maximum and minimum span evaluations.
Minimum spans are detected using both \mina and Collins' head finding rules using gold parse trees.

The corresponding results using gold mentions (system used gold mentions to resolve coreference relation), are given in Table~\ref{tab:gold_min_span}.
\begin{table*}[htbp]
    \begin{center}
	   \begin{tabular}{@{}l|lll||lll@{}}
	  \multicolumn{1}{c}{} & \multicolumn{3}{c}{CoNLL} & \multicolumn{3}{c}{LEA}\\  
	  \multicolumn{1}{c}{} & max & \mina & head & max & \mina & head \\
    \hline
    fernandes & $69.4$ & $69.4$ & $69.8$ & $56.1$  & $56.1$  & $56.1$\\
    bjorkelund& $68.0$ & $68.0$ & $68.1$ & $61.1$  & $61.1$  & $61.1$\\
    chang & $77.2$ & $77.2$ & $77.2$ & $67.9$  & $67.9$  & $67.6$\\
    chen&  $71.3$ & $71.3$  & $71.4$ & $63.9$  & $63.9$ & $63.9$\\
    yuan & $70.4$ & $70.4$ & $70.4$ & $63.4$  & $63.4$   & $63.4$ \\
    xu  & $61.0$ & $61.0$ & $61.2$ & $56.9$  & $56.9$  & $57.1$\\

    \end{tabular}
    \end{center}
    \caption{CoNLL-2012 shared task systems evaluations using gold mentions.}
    \label{tab:gold_min_span}
\end{table*}

Based on the results of Tables~\ref{conll-2012-min-max-evaluations} and ~\ref{tab:gold_min_span}: (1) the use of minimum spans reduces the gap between the performance on gold vs.\ system mentions by about two percent, (2) the use of minimum instead of maximum spans results in a different ordering for some of the coreference resolvers, and (3) when gold mentions are used, there are no boundary detection errors, and consequently the results using \mina are the same as those of using maximum spans. Due to recognizing the same head for distinct overlapping mentions, the scores using the head of gold mentions are not the same as using their maximum span, which in turn indicates \mina is suited better for detecting minimum spans compared to head words.

\end{document}